\documentclass{ecai} 

\usepackage{latexsym}
\usepackage{amssymb}
\usepackage{algorithm2e}
\usepackage{amsmath}
\usepackage{booktabs}
\usepackage{enumitem}
\usepackage{graphicx}
\usepackage{color}
\usepackage{todonotes}
\usepackage{multirow}

\newtheorem{definition}{Definition}

\newcommand{\BibTeX}{B\kern-.05em{\sc i\kern-.025em b}\kern-.08em\TeX}

\usepackage[symbol]{footmisc} 

\begin{document}


\begin{frontmatter}


\paperid{123} 


\title{Novelty Accommodating Multi-Agent Planning in High Fidelity Simulated Open World}


\author[A]{\fnms{James}~\snm{Chao}\footnote[\dagger]{These authors contributed equally to this work.}\thanks{Corresponding Author. Email: james.chao.civ@us.navy.mil}}
\author[B]{\fnms{Wiktor}~\snm{Piotrowski\footnotemark[\dagger]}\thanks{Corresponding Author. Email: wiktor.piotrowski@sri.com}}
\author[B,C]{\fnms{Roni}~\snm{Stern}}
\author[D]{\fnms{Héctor}~\snm{Ortiz-Peña}}
\author[A]{\fnms{Mitch}~\snm{Manzanares}}
\author[B]{\fnms{Shiwali}~\snm{Mohan}}
\author[A]{\fnms{Douglas}~\snm{Lange}}

\address[A]{Naval Information Warfare Center Pacific}
\address[B]{SRI International}
\address[C]{Ben Gurion University}
\address[D]{Calspan-University of Buffalo Research Center}


\begin{abstract}

Autonomous agents operating within real-world environments often rely on automated planners to ascertain optimal actions towards desired goals or the optimization of a specified objective function. Integral to these agents are common architectural components such as schedulers, tasked with determining the timing for executing planned actions, and execution engines, responsible for carrying out these scheduled actions while monitoring their outcomes.
We address the significant challenge that arises when unexpected phenomena, termed \textit{novelties}, emerge within the environment, altering its fundamental characteristics, composition, and dynamics. This challenge is inherent in all deployed real-world applications and may manifest suddenly and without prior notice or explanation. The introduction of novelties into the environment can lead to inaccuracies within the planner's internal model, rendering previously generated plans obsolete.
Recent research introduced agent design aimed at detecting and adapting to such novelties. However, these designs lack consideration for action scheduling in continuous time-space, coordination of concurrent actions by multiple agents, or memory-based novelty accommodation. Additionally, the application has been primarily demonstrated in lower fidelity environments.
In our study, we propose a general purpose AI agent framework designed to detect, characterize, and adapt to novelties in highly noisy, complex, and stochastic environments that support concurrent actions and external scheduling. We showcase the efficacy of our agent through experimentation within a high-fidelity simulator for realistic military scenarios.
\end{abstract}

\end{frontmatter}


\section{Introduction}
Current Artificial Intelligence systems excel in narrow-scoped closed worlds such as board games and image classification. However, the vast majority of realistic applications are open-world scenarios susceptible to experiencing novelties. Novelty, defined as an unknown domain shift dynamically altering the features, behavior, or characteristics of an agent or system \cite{langley2020open,kejriwal2024challenges}, poses a significant challenge for AI agents in their current state, as they prove exceedingly brittle against such novelties \cite{witty2018measuring}. This fragility results in a dramatic performance drop when operating even under slightly altered conditions \cite{chao2020evaluating}. Consequently, AI agents intended for deployment in complex realistic environments, potentially affected by unknown novelties, must possess the capacity to reason about the effects and causes of these domain shifts. Nowhere is this need more critical than in military scenarios, where the ability to comprehend and respond to novelty is paramount for protecting lives and achieving mission objectives. Resilient AI agents, operating in open worlds, must exhibit the capability to recover nominal performance in the presence of novelties through robustness and adaptability. Recently, AI Planning-based agents have showcased their formidable abilities for operating within novelty-affected open worlds \cite{piotrowski2023learning,musliner2021openmind}.

Automated planning has long been integral to military applications, with notable examples including aircrew decision aiding in modern military air missions \cite{merkel1991automated}, generating complex battle plans for military tactical forces \cite{kewley2002computational},crisis management and disaster relief \cite{wilkins1993applying}, and controlling autonomous unmanned aerial vehicles in beyond-visual-range combat \cite{floyd2017goal}. The military necessitates not only swift and decisive courses of action but also flexibility to address unforeseen circumstances. Effective crisis management is characterized by rapid response, decisive action, and adaptability to evolving environments \cite{wilkins1993applying}.

Currently, the most prevalent methods for addressing planning failures encompass replanning and plan repair~\cite{fox2006plan,van2005plan}. These methods can handle environmental uncertainty to a limited extent. However, they assume that the autonomous agent's internal model of its environment is still correct and accurate beyond a minor discrepancy in plan execution or perception. In real-world scenarios, novelty can significantly alter the fundamental characteristics and dynamics of the environment. In such instances, replanning and plan repair methods fail repeatedly, persistently utilizing an outdated and inaccurate world model. Effectively adapting to the updated environment necessitates understanding the changes and repairing the agent's internal model to accurately reflect the unknown domain shifts.


In this work, we adapt Hypothesis-Guided Model Revision over Multiple Aligned Representations (HYDRA) \cite{mohan2023domain}, an established AI planning-based novelty-aware framework for designing intelligent agents, for executing complex real-world scenarios in high- and medium-fidelity military simulators. Previously, HYDRA has found application in single-agent environments of lower fidelity, such as physics-based games like Angry Birds \cite{9619160} and OpenAI Gym's CartPole \cite{gym}. Our contribution  extends HYDRA's applicability to multi-agent environments characterized by high levels of noise, stochasticity, and complexity, a domain previously unexplored. The integration of novelty-aware functionality is particularly pertinent in military contexts, where the accurate representation and understanding of environmental phenomena and adversary behavior are paramount. We substantiate our approach by demonstrating enhancements in AI agent decision-making performance compared to baseline models.

\section{Related Work} \label{rw}
Several theories of novelty have been proposed in recent literature \cite{boult2021towards,langley2020open,alspector2021representation,doctor2022toward}.  Generally, novelty refers to an unknown and unexpected domain shift that alters a given system, its dynamics, or the entities operating within the system. One particularly relevant explanation of novelty, as it pertains to our work, defines it as spatio-temporal, structural, process, or constraint environmental transformations \cite{langley2020open}. More recently, \cite{doctor2022toward} proposed a theory that distinguished eight separate novelty classes. We adopt the definition provided by \citet{chadwick}, who maps Doctor's theory and its novelty classes to the military domain.

Research on novelty has been conducted across various domains, including Monopoly \cite{KEJRIWAL2021102364}, Minecraft~\cite{musliner2021openmind,muhammad2021novelty}, Angry Birds~\cite{9619160}, Doom~\cite{holder}, CartPole~\cite{boult2021towards}, natural language processing~\cite{ma-etal-2021-semantic}, and computer vision~\cite{Girish_2021_ICCV}. Work on novelty-accommodating approaches within these domains has also been undertaken, employing methodologies such as probabilistic reasoning~\cite{loyall2022integrated}, knowledge graphs~\cite{balloch2023neuro}, causal models~\cite{jensen2021improving}, belief validation~\cite{muhammad2021novelty}, change-point detection~\cite{musliner2021openmind}, image novelty accommodation~\cite{du2023novel,pang2022andea}.

HYDRA~\cite{mohan2023domain} stands as a domain-independent AI framework utilizing a model-based planning approach to detect, characterize, and adapt to various classes of unknown novelties. Its efficacy has been demonstrated across multiple domains, including Angry Birds, Minecraft, and CartPole3D. To describe the agent's environment and its dynamics, HYDRA employs PDDL+ \cite{fox2006modelling}, a standardized planning modeling language for mixed discrete-continuous systems. PDDL+ is highly expressive, facilitating accurate and comprehensive capturing of the structure and dynamics of the target system as a planning domain. 


HYDRA streamlines the process by automatically generating PDDL+ problem instances from environmental perception data, combining them with a manually crafted domain to create a comprehensive planning model. This domain only requires one-time composition per environment. Using a domain-independent PDDL+ planner, HYDRA solves the planning task and executes the resulting actions in the simulation environment. After execution, HYDRA employs the PDDL+ planning model to identify novelty through inconsistency checking. These checkers detect deviations between expected behavior (the planning trace) and observed data during execution. They can be general, considering all PDDL+ state variables, or focused, targeting specific subsets of variables to reason about phenomena like unexpected entity behavior or action effects.

Initially, the default PDDL+ planning model solely accounts for nominal, non-novel behavior. 
Once a significant novelty is detected through inconsistency checking, HYDRA invokes the repair module to find an explanation for the divergent behavior. The PDDL+ planning model is iteratively modified, in an attempt to incorporate the explanation of the novelty into the planning model, until the planning-based predictions realign with observations and performance is recovered. The planning model repair is performed via a heuristic search process that adjusts the values of model variables to minimize inconsistency (i.e., the euclidean distance between expected and observed state trajectories).

To operationalize HYDRA's novelty detection, characterization, and accommodation capabilities, a centralized planning approach for multiple agents has been developed, integrating components including PDDL+, heuristic functions, a scheduler, and an execution engine within the high and medium fidelity simulator. Comprehensive details regarding this design are provided in \cite{chaotop}. 


HYDRA is compared against baseline agent approaches, namely the default domain-specific planners integrated into the high and medium fidelity simulation environments: High fidelity being Simulation Orchestration for Learning and Validation Environment (SOLVE) \cite{chadwick} paired with Advanced Framework for Simulation, Integration and Modeling (AFSIM) \cite{afsim}, while the medium fidelity simulator is an internally developed variant of AFSIM with reduced elements of compounding stochasticity effects. The baseline agents exhibit complete failure when encountering novelty, underscoring the necessity of applying adaptive methods, such as HYDRA, to ensure expected behavior and performance in the presence of unexpected environment changes.

\section{Problem Statement} \label{section:pf}

The simulated mission sees military aircraft being deployed to destroy ground enemy targets, while avoiding being hit by surface-to-air missiles (SAMs) fired by the enemy. Throughout the mission, we anticipate any novelty where unexpected phenomena alter the environment in unknown ways violating the assumptions of our planning AI agent. The HYDRA agent does not have any a priori knowledge about the novelty, its severity, time of occurrence, or characteristics. This results in challenges that surpass the capabilities of current state-of-the-art methods to recover performance in a timely manner.

\begin{figure}[h]
    \centering
    \includegraphics[width=8.5cm, trim={1cm 3cm 3cm 4cm},clip]{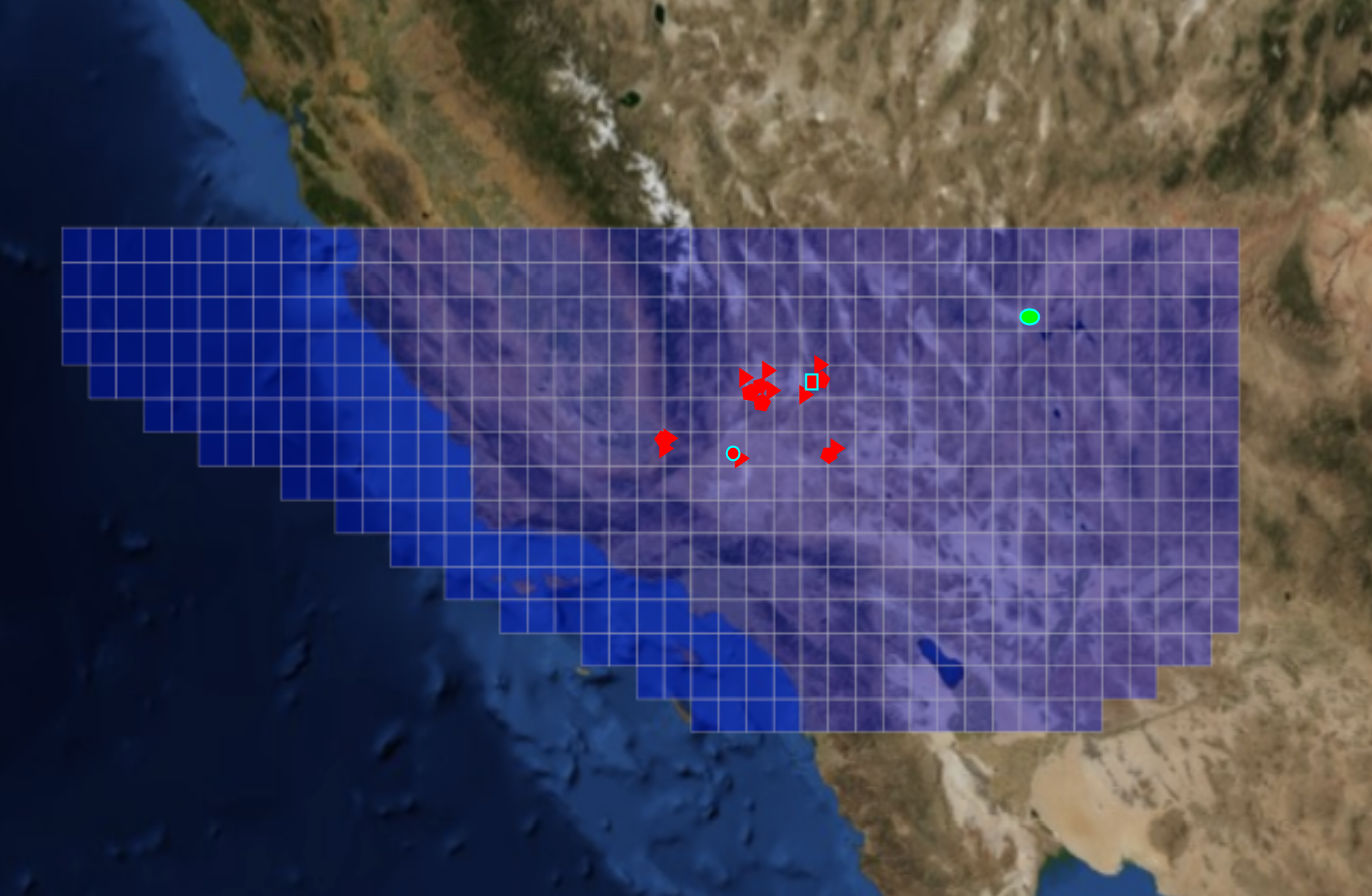}
    \caption{Mission Area Grid And Entities: teal circle is the aircraft and autonomous surveillance drone, red rectangle with teal outline is the enemy SAM, red circle with teal outline is the target radar station, red triangles are enemy radar sensors, and red pentagons are all other enemy entities}
    \label{fig:grid}
\end{figure}

The observations of enemy entities for both experiments will be assumed to be gathered by the surveillance drone at the onset of all engagements. The mission area encompasses various elements including 10 enemy radars utilized by different military commands within the enemy chain of command, a supply depot, an ammunition storage station, a surface-to-air missile launcher (SAM), a chemical storage unit, a command post, and a defense headquarters. The mission area and entities are shown in Figure \ref{fig:grid}. The teal circle denotes the initial position of the F-35 aircraft and also serves as the starting location for the autonomous surveillance drone. The red rectangle outlined in teal indicates the location of the enemy SAM site, while the red circle with a teal outline signifies the position of the target radar station (primary objective). Additionally, red triangles represent enemy radar sensors, and red pentagons denote other enemy entities within the area.

\section{Novelty Accommodating Agent}

\subsection{Planning Domain Formalization}
Novelty detection, characterization, and accommodation are the primary focus of this work, a formal definition of the novelty response problem is essential. Building upon the definition of a planning domain and the concept of novelty, we employ the HYDRA methodology \cite{mohan2023domain} to quantify the degree of inconsistency between the planning model and the environment, which in our case is an open real-world environment represented by high and medium fidelity simulators. Finally, we establish a set of model manipulation operators (MMOs) to facilitate meta-model repair until the model aligns more closely with the environment.
Formally: 
\begin{definition} \textbf{Environment}
Let E be the environment, a transition system defined as:
\begin{equation}
E = \langle \mathcal{AG}, \mathcal{S}, \mathcal{S_I}, \mathcal{A}, \mathcal{E}, \mathcal{G}, \mathcal{V} \rangle
\end{equation}
where $\mathcal{AG}$ is a set of agents, $\mathcal{S}$ is the infinite set of states, $\mathcal{S_I} \subseteq \mathcal{S}$ is a set of possible initial states, $\mathcal{A}$ is a set of possible actions, $\mathcal{E}$ is a set of possible events, $\mathcal{G}$ is a set of possible goals, and $\mathcal{V}$ is a set of domain variables including both proposition and numeric.
\end{definition}

\begin{definition} \textbf{Novelty response problem}
Let $\prod$ be the novelty response problem \cite{stern2022model}  defined as: 
\begin{equation}
\prod = \langle E, \varphi , t_N \rangle
\end{equation}
where E is the transition system environment, $\varphi$ is the novelty function, and $t_N$ is a non-negative integer specifying the battle in which novelty $\varphi$ is introduced within a tournament. 
\end{definition}
In the military domain we operate in, the terms battle is interchangeable with episode, and campaign is interchangeable with tournament.



\begin{definition} \textbf{Inconsistency}
Let C be the inconsistency checking function defined as:
\begin{equation} \label{cequ}
C(\mathcal{M}, E, \pi, t_e, t_o) \rightarrow \mathbb{R}_{\geq 0}
\end{equation}

where $\mathcal{M}$ is the PDDL+ model of and internal approximating model of environment E, $\pi$ is a sequential plan that solves E to reach $\mathcal{G}$, $t_e$ is a trajectory we expect to observe using model $\mathcal{M}$ to solve the planning problem, and $t_o$ is the trajectory we actually observe in the environment when applying the plan $\pi$. 
\end{definition}

A trajectory is a sequence of state-agent-action tuples, generated by executing a plan in an environment. $t_e$ in the PDDL+ model as shown in equation \ref{eq:te}. $t_o$ in the real environment E such as the high and medium fidelity simulators as shown in equation \ref{eq:to}.

\begin{equation} \label{eq:te}
    t_e(\mathcal{M}, \pi) = (\langle s_0, \pi_1, a_1 \rangle, ..., \langle s_n, \pi_n, a_n \rangle)
\end{equation}
\begin{equation} \label{eq:to}
    t_o(E, \pi) = (\langle s_0, \pi_1, a_1 \rangle, ..., \langle s_n, \pi_n, a_n \rangle)
\end{equation}




\begin{definition} \textbf{MMO} A model manipulation operator (MMO) is a single change to the agent's internal model. Specifically, we limit the scope of MMOs to modifying the values of variables present in the agent's internal model by a predetermined interval. MMOs are then utilized within the model repair mechanism to accommodate novelty by adapting the agent's internal model. Essentially, the repair process aims to identify a sequence of MMOs that, when applied to the agent's internal model $\mathcal{M}$ yields an updated model $\mathcal{M'}$ which accounts for the introduced novelty. In the scope of this work, an MMO is a function $m : V \times \Delta V \rightarrow V$, where $V \in \mathbb{R}$ is a numeric or boolean variable present in the agent's model $\mathcal{M}$, and $\Delta V \in \mathbb{R}$ is the numeric or Boolean change to the value by an interval treated as a parameter. 
\end{definition}

In practice, an MMO is applied in a straightforward manner $v(\mathcal{M}) = v(\mathcal{M}) \pm \Delta v(\mathcal{M})$ where $v(\mathcal{M})$ is the numeric value of some variable in model $\mathcal{M}$ and $\Delta v(\mathcal{M})$ is a predefined change interval specific to that model variable $v(\mathcal{M})$. This approach can also be extended to propositions $p(\mathcal{M})$ by casting each one as a numeric variable $v(p(\mathcal{M}))$. After MMOs have been applied, the variable is then re-cast into a true proposition such that $p(\mathcal{M}) {=} True$ if $v(p(\mathcal{M})) {>} 0$ and $p(\mathcal{M}) {=} False$ otherwise. 

Based on the above MMO definition, a repair $R$ is a function which takes in a model $\mathcal{M}$ and a sequence of MMOs $\{m\}$ that modify the model's variables, the repair returns the modified model $\mathcal{M}'$. 

\begin{definition} \textbf{Repair}
    Let repair $R$ be a function $R(\mathcal{M}, \{m\}) \rightarrow \mathcal{M'}$, where $\mathcal{M}$ is the model of the environment E, and $\{m\}$ is the a set of MMOs defined over model $\mathcal{M}$. The repair function yields an updated model $\mathcal{M}'$ which is generated by applying $\{m\}$ to the default model $\mathcal{M}$.
\end{definition}

$R$ is determined to be a useful repair if it yields a smaller inconsistency $C$ according to equation \ref{cequ}.



\subsection{Agent Architecture}
\begin{figure}[h]
    \centering
    \includegraphics[width=8cm]{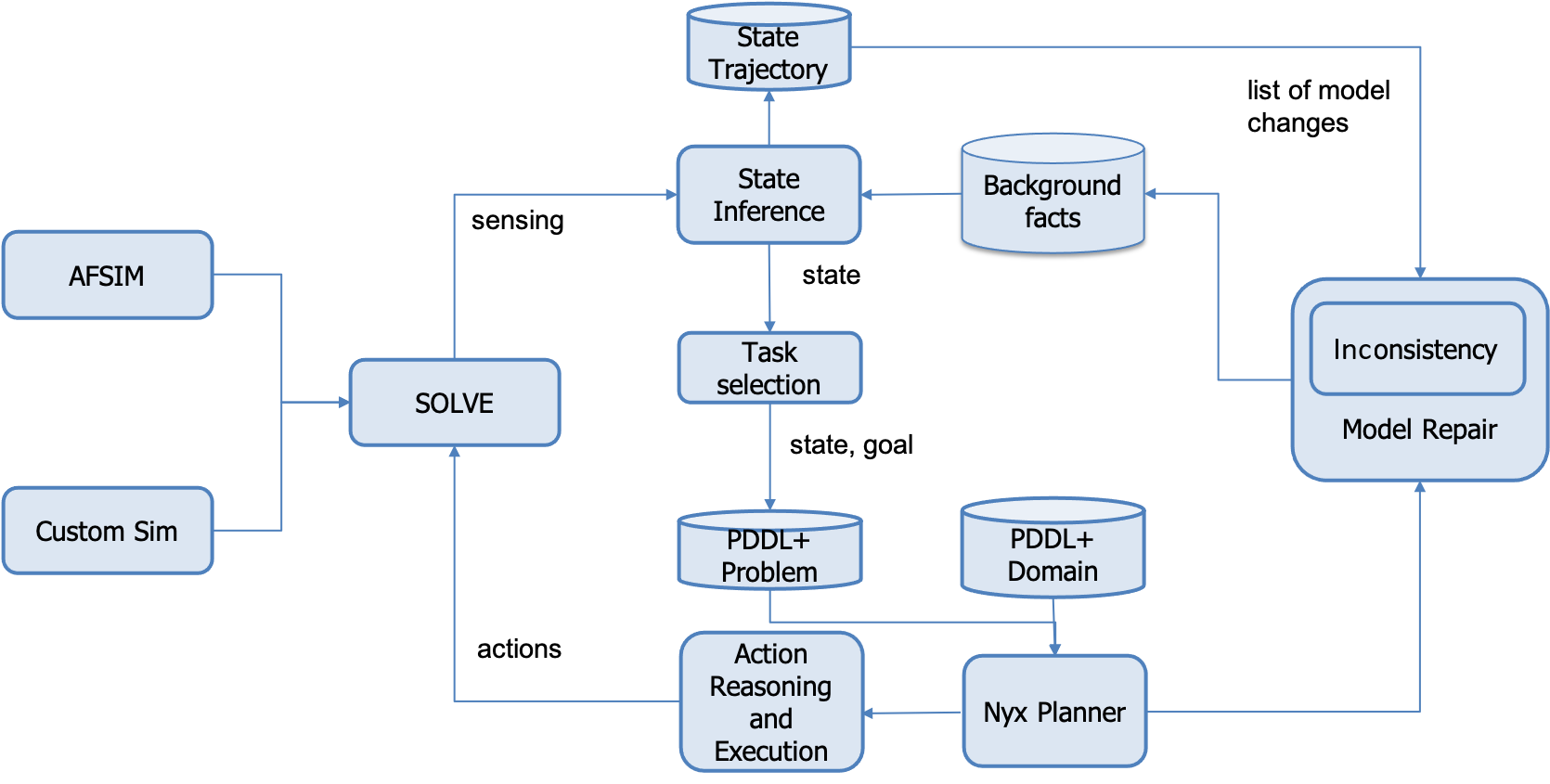}
    \caption{Agent Architecture}
    \label{fig:comp}
\end{figure}

Figure \ref{fig:comp} describes the components and information flow of the agent architecture. Initially, a PDDL+ problem file is automatically generated from initial observations and intrinsic assumptions about the environment. Subsequently, a full model $\mathcal{M}$ is created by combining the auto-generated problem with a general manually-defined PDDL+ domain. A novel domain-independent PDDL+ planner, Nyx~\cite{piotrowski2024realworld}, then uses the model $\mathcal{M}$ to solve for a plan $\pi$, an execution engine built into SOLVE translates the plan into low-level instructions executable in the high- and medium-fidelity simulation environments. Following each battle $t_N$ in the simulator, the agent calculates an inconsistency score $C$ comparing the expected outcomes $t_e$ and observed outcomes $t_o$. If inconsistency $C$ exceeds threshold $\mathcal{T}$ it is likely that the underlying environment $E$ has been substantially altered by novelty, beyond interference from noisy sensor readings or expected stochasticity. In response, the agent initiates the meta-model repair process, which adjusts the model $\mathcal{M}$ by iteratively applying MMOs $m$. The repair process updates the background facts of the the PDDL+ problem file such that an explanation of the detected novelty is incorporated into the agent's internal reasoning model.

\subsection{Domain Knowledge}

PDDL+ serves as a domain-independent method for capturing essential functional knowledge about the environment and its agents. PDDL+ is expressive and general enough to facilitate modeling and solving of vastly diverse systems, such as Angry Birds~\cite{piotrowski2023heuristic}, Urban Traffic Management~\cite{vallati2016efficient}, UAV routing~\cite{kiam2020ai}, or military missions described in this paper. PDDL+ models must mitigate scalability and complexity issues while maintaining accuracy of characteristic features and behavior of the modeled system. The domain knowledge contained within the PDDL+ model for the experiments conducted in this paper encompasses various aspects, including the behavior of enemy and neutral entities, characteristics of enemy weapon systems, environmental phenomena, temporal evolution of the world. MMOs, HYDRA's atomic model changes that comprise a repair, further demonstrate domain knowledge. In practice, only impactful aspects of the planning model will be modified. Thus, selecting the right subset of model elements is crucial for HYDRA's efficiency. Domain knowledge encoded in PDDL+ is indispensable for the application of HYDRA to any new problem scenario. It is crucial for detecting and accommodating various novelties in the open real world within a reasonable timeframe.

\subsection{Inconsistency Checking}
The inconsistency measure $C \in \mathbb{R}_{\geq 0}$ is a calculated number representing how accurate the model $\mathcal{M}$ represents the high and medium fidelity simulator transition system $E$. Inconsistency $C=0$ means that the agent is operating on an internal model $\mathcal{M}$ that is perfectly aligned with the simulation environment $E$. The higher the value of $C$, the less accurate the agent internal model represents the environment E. $t_e$ is simulating using the PDDL+ model $\mathcal{M}$. $t_o$ is generated by converting direct observations of the environment into PDDL+ to calculate $C$ using equation \ref{eq:C} (because $t_e$ and $t_o$ has to be the same datatype). While comparing every single variable $V$ will be ideal in catching any model inconsistency, however, computation limitations of calculating $\infty$ amounts of $\Delta V$ makes that infeasible, so we focus on comparing the Euclidean distance between selected important state variables that represent the system dynamics enough to result in inconsistency which effects overall performance. The state variables do not map one to one to unknown novelty, but rather to parameters an agent has at its disposal to detect and accommodate the novelty. 

\begin{equation} \label{eq:C}
    C = \sum_{ag} \sum_i  \gamma^i \cdot \| t_o(\pi, E)[ag][i] - t_e(\pi, \mathcal{M})[ag][i] \|
\end{equation}
where $t_o(\pi, E)$ is the observed trajectory of executing the plan $\pi$ in the environment E. And $t_e(\pi, \mathcal{M})$ is the expected trajectory of executing the plan $\pi$ using the PDDL+ model. $0 < \gamma < 1$ is a discount factor to account for compounding errors of Euclidean distances for later states. $i$ denotes the current step that exists in the transition system $E$. $ag$ denotes a specific agent in the multi-agent setup.

However, in the application to real world problems, we modify the inconsistency checker by setting $i$ to always equal the last observed (terminal state) and expected states (goal state). This is because the original HYDRA system was designed in a single agent discrete-time environment, where the plan $\pi$ generated by the PDDL+ can be executed on a transition system $E$ as shown in equation \ref{pddlcc}. This generates the expected trajectories or states $t_e$ where $\{ s_0, s_1, ..., s_n  \}$ denotes a specific state before an agent applies an action, $\{ \pi_0, \pi_1, ..., \pi_n \}$ denotes the plan for a specific agent in $\mathcal{AG}$, and $\{ a_0, a_1, ..., a_n \}$ denotes a specific action of a specific agent depending on the preceding agent within $\{ \pi_0, \pi_1, ..., \pi_n \}$.

\begin{equation} \label{pddlcc}
    s_0 \pi_1 a_1 \rightarrow s_1 \pi_1 a_2 \rightarrow s_2 \pi_1 a_3 \rightarrow s_3 \pi_2 a_1
\end{equation}

When transitioning a PDDL+ plan to a multi-agent setting in a high or medium fidelity simulator, the solution is to query the state space which creates a transition system $E$ shown in equation \ref{macc}. This becomes the observed trajectories or states $t_o$ with the same variable definitions as the expected states in equation \ref{pddlcc}. There is no guarantee $\pi_1$ will not execute multiple actions before $\pi_2$ executes its first action in the continuous multi-agent high and medium fidelity simulators. Resulting in mismatches between the states, as $s1$ from equation \ref{macc} will not exist in equation \ref{pddlcc}.

\begin{equation} \label{macc}
s_0 
\begin{cases}
\pi_1 a_1 \\
\pi_2 a_1 \\
...
\end{cases}
 \rightarrow
s_1 
\begin{cases}
\pi_1 a_2 \\
\pi_2 a_2 \\
...
\end{cases}
 \rightarrow
s_2 
\begin{cases}
\pi_1 a_3 \\
\pi_2 a_3 \\
...
\end{cases}
 \rightarrow
s_3 
\begin{cases}
\pi_1 a_4 \\
\pi_2 a_4 \\
...
\end{cases}
\end{equation}

Furthermore, the high and medium fidelity simulator exhibit sufficient noise levels such that not every action or event is executed at a defined time step. Consequently, the transition system $E$ is represented as shown in Equation \ref{noisycc}, wherein no action can occur for several time steps.

\begin{equation} \label{noisycc}
s_0 
\begin{cases}
\pi_1 a_1 \\
\pi_2 a_1 \\
...
\end{cases}
 \rightarrow
s_0
\begin{cases}
\pi_1 a_1 \\
\pi_2 a_1 \\
...
\end{cases}
 \rightarrow
s_0
\begin{cases}
\pi_1 a_1 \\
\pi_2 a_1 \\
...
\end{cases}
 \rightarrow
s_1 
\begin{cases}
\pi_1 a_2 \\
\pi_2 a_2 \\
...
\end{cases}
\end{equation}

In theory, the issue could be addressed by conducting a search to identify identical states. However, in the high and medium fidelity simulator, the state transition time is continuous, leading to the formulation of transition system $E$ as shown in Equation \ref{contcc}. In this context, a state $s$ existing in the observed space $t_o$ is never precisely identical to a state $s \in$ expected states $t_e$. Even if states were approximated within a time interval to closely match the expected states, distinguishing between measurement errors and novelty effects would remain challenging. Moreover, discerning whether novelty arises as a consequence of concurrent actions across multiple agents is difficult. Additionally, the absence of a concept regarding when an action is completed, as all actions are scheduled at the outset without environment feedback, poses challenges for the agent to accurately determine when to create a state $s \in$ $\mathcal{S}$ post-action $a \in$ $\mathcal{A}$ for comparison against expected states from the PDDL+ plan.

Novelty in the military domain can be introduced through news or post-battle reports, which are not inherently part of the observation. Therefore, inconsistency checking must incorporate the outcomes of such reports. HYDRA compares expected results with observations, and thus, the reports will be appended to the observations space as the final observation, forming part of the terminal state.

% ...

 \begin{equation} \label{contcc}
 \scriptsize
s_{[0-1]} 
\begin{cases}
\pi_1 a_1 \\
\pi_2 a_1 \\
...
\end{cases}
 \rightarrow
s_{[0-1]} 
\begin{cases}
\pi_1 a_1 \\
\pi_2 a_1 \\
...
\end{cases}
 \rightarrow
s_{[1-2]} 
\begin{cases}
\pi_1 a_2 \\
\pi_2 a_1 \\
...
\end{cases}
 \rightarrow 
s_{[1-2]}  
\begin{cases}
\pi_1 a_3 \\
\pi_2 a_2 \\
...
\end{cases}
\end{equation}

% ...

Each difference between $t_e$ and $t_o$ can be scaled by the importance. For example, a friendly aircraft that is unexpectedly destroyed has a higher importance weight than inaccuracies in the hit points (HP) of a target. In the presented aircraft mission scenario, the inconsistency score is specifically weighted to encourage the agent to prioritize repairing the model towards keeping friendly aircraft alive. Other novelties are weighted via inconsistency shaping to help agents prioritize repairs.

Environmental zones model severe inclement weather that exhibit highly stochastic behavior where we cannot predict an outcome. For example, an aircraft can spin in a tornado or simply move to an unexpected location. And no two results are consistent once entering environmental zones, as state variables such as location and speed of the aircraft are constantly changing, and impossible to reason if it is a novelty or just an anomaly. To accommodate this, we apply a weak fault \cite{khalastchi2019fault} strategy of inconsistency checking, we consider a parameter $\Delta_{x,y}$ that is the Euclidean distance between the expected location and the observed location at the end of the battle. If $\Delta_{x,y}$ $>$ a threshold learned from data of historical results, then a PDDL+ predicate will indicate that we failed to move due to novelty.

\subsection{Meta Model Repair}
The meta model repair is triggered once the inconsistency $C$ of the default model $\mathcal{M}$ exceeds the threshold $\mathcal{T}$. Repair aims to alter the model $\mathcal{M}$ so that the expected trajectory $t_e$ (yielded by the planner, based on $\mathcal{M}$) is consistent with the trajectory $t_o$ observed when executing plan $\pi$ in simulation environment $E$. As stated in \cite{stern2022model}, defining appropriate MMOs is the key to a good repair that leads to good novelty accommodation. Although monitoring every variable $v \in \mathcal{M}$ will create the most general agent, the search space will become too large to find feasible repairs in a reasonable time. Thus, currently, model variables which MMOs will modify during repair are chosen manually. In the presented problem dynamics, the MMOs we labeled are: 1) An area of environment zones 2) An area of no-fly-zones 3) If a no-fire-entity is within collateral damage distance to a target 4) Each enemy entity missile range 5) An enemy entity HP.


The state variables in the observation space we focus on to reduce complexity: 1) Each friendly aircraft status 2) Each enemy entity status 3) Each friendly aircraft location 4) A news report of no-fire-entity destroyed 5) Each neutral entity status.


For object level novelty such as no-fire-entities. Once we are told novelty exists via a report, the repair will simulate the MMO of no-fire-entity near a target, for all existing targets. Until we find a repair that matches the fact that we destroyed a no-fire entity.

Regarding relation level novelty such as increased survivability of the target, the MMO modeling the HP of the SAM are repaired. Often repairing the HP of the SAM will result in increased inconsistency, so we limit the ability to repair the SAM to 0 or introduce a PDDL+ predicate to limit the fire abilities upon novelty that causes such phenomenon or track each aircraft on if they fired their missiles.

In the case of environment level novelties such as severe weather, the repair will check every possible combination of locations of the grid, causing computation to exponentially increase. To accommodate for this issue, we extended HYDRA with: 1) Introduce memory to the repairs, to make sure of any repaired zones. Remembered repairs will be denoted as $R_r$. 2) Introduce non-relative repairs to the current HYDRA meta model, so that repairs can encode visited zone repairs. 3) Exclude initial state as an environment zone as that will result in an aborted mission.


The search-based algorithm for model repair is domain-independent. The model repair will follow the algorithm \ref{alg:env}, which shows the pseudo-code for the meta model repair. First, we check the inconsistency score $C_{best}$, if $C_{best} > \mathcal{T}$ then we apply MMOs \{m\} one by one. For each MMO, we check the inconsistency $C_{new}$, if any $C_{new}$ is less than the best inconsistency score $C_{best}$, then the agent will deem the MMO repair as successful and update the model $\mathcal{M}$. This process is repeated until the smallest $C_{best}$ is found.



In our extended version of the repair algorithm designed to address novelty correlated with geographic locations, instead of generating and simulating repairs without considering the relative positioning of the grid cells, the agent now remembers repair locations in sequence. For instance, if the best repair involves updating the model zone location (0,1) relative to the agent's location, then in the subsequent battle, the agent will avoid choosing the repair relative to its current position (0,-1) as the best option, as it would lead back to the already-considered initial location. Each battle, a new repair $R$ is added to the $R_r$ list, and the newly added repair is recorded by a relative path based on the last repair position.



\RestyleAlgo{ruled}
\begin{algorithm}
\caption{Repairing Environment Novelty}
\label{alg:env}
$C_{best}$ $\gets$ EstimateInconsistency($\mathcal{M}$, E, $\pi$, $t_e$, $t_o$, $\mathcal{T}$)\\
\While{$C_{best}$ $\geq$ $\mathcal{T}$}{
    \ForAll{$MMO$ $\in$ $m$}{
        $\mathcal{M'} \gets R(\mathcal{M}, MMO)$ \\ 
        $C_{new}$ $\gets$ EstimateInconsistency($\mathcal{M}$, E, $\pi$, $t_e$, $t_o$, $\mathcal{T}$)\\
        \If{$C_{new} < C_{best}$  $\land$ $R$ $\notin$ $R_{r}$}{
            $C_{best}$ $\gets$ $C_{new}$
            \If{$C_{best} \leq \mathcal{T}$}{
                $\mathcal{M}$ $\gets$ $\mathcal{M'}$
            }
        }
        UndoUpdate$\mathcal{M}$($MMO$)
    }
    
    \If{$R$ $\notin$ $R_{r}$}{
        \ForAll{$R_{r}$}
        {$R_r[i]$ = $R_r[i]$ - $R$}
        $R_{r}$.append($R$) \\
    }
           
    return $\mathcal{M}$
}
\end{algorithm}

            
        

Finally, a heuristic is used to guide the repair process to prioritize: 1) Repairs that are more consistent with observations. 2) Repairs that focus on significantly updating a single model aspect over complex explanations that shallowly adjust multiple different model elements.


The repair process is a search task which iteratively considers different changes to the model via pre-defined MMOs (e.g., increase weapon range by 1 cell,  increase weapon range by 2 cells, decrease weapon range by 1 cell, etc.). The goal of this search is to sufficiently reduce the inconsistency between expected and observed state trajectories. The process terminates once the inconsistency score falls below the predefined threshold.

    

\begin{table*} [h]
\centering
\caption{Inconsistency and Repair scores for HYDRA agents in medium fidelity simulator} 
\begin{tabular}{|c|c|c|c|c|c|c|}
\hline
    \multirow{2.4}{*}{\textbf{Novelty}} & \multirow{1.3}{*}{\textbf{Inconsistency}} & \multirow{1.3}{*}{\textbf{Repair}} & \multicolumn{2}{c|}{\multirow{1.2}{*}{\textbf{Repair Performance (\% Inconsistency reduction)}}} & \multirow{1.2}{*}{\textbf{Generated Repair}} & \multirow{2}{*}{\textbf{\# Repairs}} \\ 
     & \textbf{Score} & \textbf{Time (s)} & \textbf{\hspace{0.4cm} Current Battle \hspace{0.4cm}} & \textbf{Next Battle} & \textbf{(Change to Model)} & \\ \hline
    Object & 1 & 30.55 & 100\% & 100\% & No Fire Entity + 1 & 1\\ \hline
    Agent & 20 & 74.02 & 95\% & 100\% & Red Weapon Range + 1 & 1\\ \hline
    Relation & 1 & 6.51 & 100\% & 100\% & Enemy target HP + 1 & 1\\ \hline
    Environment & 10 & 51.39 & 100\% & 100\% & Zone Row - 2, Zone Col - 1 & 3\\ \hline
    Goal & 20 & 31.17 & 95\% & 100\% & Red Weapon Range + 1 & 1\\ \hline
    Event & 20 & 31.17 & 95\% & 100\% & Red Weapon Range + 1 & 1\\ \hline
    
\end{tabular}
\label{tab:consrepair} 
\end{table*}


\section{Experiment Setup}
Experimental evaluation was conducted on a machine with MacOS, an Intel Core i7 2.6GHz 6-core CPU with 16GB DDR4 memory.

\subsection{Agents}
Our evaluation compares HYDRA with a baseline agent. The baseline is agent is identical to HYDRA but does not perform novelty detection or accommodation (repair). The baseline agent persistently uses the default non-novel PDDL+ model to act in the environment whereas the HYDRA agent updates the PDDL+ model based on the observations and expectations. Thus, the comparison focuses on the impact of novelty reasoning and repair on overall agent performance and mission success. 

\subsection{Novelties}
Examples of military-related novelties, categorized within the eight categories outlined in \cite{chadwick}, are provided below. These examples serve to illustrate the potential impact and scope on agent performance, and their relevance in real-world operations. It's important to note that novelties encountered in high and medium fidelity simulators extend beyond these instances. Many historical incidents involve novelty, the U.S. airstrike aimed at terrorists in Kabul that killed additional people, causing human rights groups to worry~\cite{hashimy2023deployment} is an example of object novelty. A U-2 spy plane shot down by Soviet air defense while believed to be beyond the reach of enemy missiles is an example of agent novelty~\cite{powers2004operation}. The world's first steam-powered ironclad warship~\cite{coski2012css} is an example of relation novelty due to unexpectedly high survivability. A rescue helicopter caught in a sandstorm trying to rescue captive embassy staff~\cite{russell1947crisis} is an example of environment novelty. 
Object novelty: No fire entities. Agent novelty: Range of an enemy SAM increases. Action novelty: SAM site becomes mobile. Relation novelties: No fire zone, unexpected target hardening. Interaction novelty: Asset communication radius change. Environment novelty: Weather affects speed of asset movement; sudden swarms of military drones. 
Goal novelty: Enemy aircraft changes patrolling area. Event novelty: New advanced enemy aircraft appears.


In the experiment, object, agent, relation, environment, goal, and event novelties were introduced. 


\subsection{Novelty Injection and Agent Performance}

A 20 battle campaign $\prod$ is conducted, where a randomly chosen novelty $\varphi$ is injected into a random battle $t{_{N}}$ . The novelty strength is sampled from a Gaussian distribution. Both a baseline non-novelty accommodating agent and a HYDRA novelty accommodating agent are tested in the campaign. Performance measure metrics for novelty detection and accommodation are detailed in paper \cite{pinto2020measuring}. 




\section{Results}
The baseline and HYDRA agent detection and accommodation performance for the high fidelity simulator, SOLVE, is shown in Table \ref{tab:hifi}. Additionally, performance for the medium fidelity simulator, AFSIM, is shown in Table \ref{tab:medfi}. Notably, the high-fidelity simulator introduces several complexities, including chained stochasticity, partial observability, dynamics, sequentiality, continuous time and space, asymmetry, high noise levels, and variance in stochasticity and novelty between battles, compared to the medium-fidelity simulator. These complexities yield intriguing outcomes such as instances where friendly aircraft are targeted by enemy SAMs but escape unscathed, enemy HP requiring more missiles than the equipped maximum for destruction, and instances where novelty does not significantly impact the agent's performance to alter the outcome. The PDDL+ model for the high-fidelity simulator is not yet fully developed to address all pre-novelty simulator stochasticity and variances compared to the baseline agent.

Table \ref{tab:consrepair} showcases repair details and performance for the medium-fidelity simulator experiments. The repair performance is measured by post-adaptation inconsistency reduction. The \textit{Current Battle} performance score denotes the gain during repair in the battle where novelty has been detected, while the \textit{Next Battle} score represents the consistency gain during the subsequent battle using the newly repaired planning model. The number of repairs indicates the battles required for repair to recover perfect inconsistency score. 


\begin{table*} [h]
\centering
\caption{Performance of baseline and HYDRA novelty accommodating agents in the high fidelity simulator.} 
\begin{tabular}{|l|l|l|l|l|l|l|l|l|}
    \hline
    Novelty Category & \multicolumn{2}{|c|}{Object} & \multicolumn{2}{|c|}{Agent} & \multicolumn{2}{|c|}{Relation} & \multicolumn{2}{|c|}{Environment}\\ \hline
    Performance Metric & Baseline & HYDRA & Baseline & HYDRA & Baseline & HYDRA & Baseline & HYDRA\\ \hline 
    False neg\% & 100 & \textbf{5} & 100 & \textbf{0} & 100 & \textbf{0} & 100 & \textbf{6}\\ \hline
	Target destroy\% post-nov & 0 & \textbf{5} & 0 & \textbf{9} & 0 & \textbf{64} & 0 & \textbf{71}\\ \hline
    Aircraft survive\% post-nov & 0 & \textbf{100} & 0 & \textbf{64} & \textbf{100} & 93 & \textbf{100} & \textbf{100}\\ \hline
    Target destroy\% pre-nov & \textbf{100} & \textbf{100} & \textbf{100} & 78 & \textbf{100} & 83 & \textbf{100} & \textbf{100}\\ \hline
    Aircraft survive\% pre-nov & \textbf{100} & \textbf{100} & \textbf{100} & \textbf{100} & \textbf{100} & \textbf{100} & \textbf{100} & \textbf{100}\\ \hline
	Detection\% & 0 & \textbf{100} & 0 & \textbf{100} & 0 & \textbf{100} & 0 & \textbf{94}\\ \hline
	False pos\% & - & 0 & - & 22 & - & 17 & - & 0\\ \hline
    No fire entity destroy post-nov & 100 & \textbf{5} & - & - & - & - & - & -\\ \hline
    No fire entity destroy pre-nov & 100 & 100 & - & - & - & - & - & -\\ \hline

\end{tabular}
\label{tab:hifi} 
\end{table*}


\begin{table*} [h]
\centering
\caption{Performance of HYDRA novelty accommodating agents in the medium fidelity simulator.} 
\begin{tabular}{|l|l|l|l|l|l|l|l|l|}
    \hline
    Novelty Category & Object & Agent & Relation & Environment & Goal & Event\\ \hline
    False neg\% & 5 & 5 & 5 & 5 & 5 & 5\\ \hline
	Target destroy\% post-nov & 5 & 100 & 100 & 85 & 100 & 100\\ \hline
    Aircraft survive\% post-nov & 100 & 50 & 100 & 100 & 50 & 50\\ \hline
    Target destroy\% pre-nov & 100 & 100 & 100 & 100 & 100 & 100\\ \hline
    Aircraft survive\% pre-nov & 100 & 100 & 100 & 100 & 100 & 100\\ \hline
	Detection\% & 100 & 100 & 100 & 100 & 100 & 100\\ \hline
	False pos\% & 0 & 0 & 0 & 0 & 0 & 0\\ \hline
    No fire entity destroy pre-nov & 5 & - & - & - & - & - \\ \hline

\end{tabular}
\label{tab:medfi} 
\end{table*}

In the context of object novelty, the agent obtains a post-battle report through a news channel. This type of novelty differs from any other encountered by HYDRA in Angry Birds, Minecraft, and CartPole, as it is not caused by an action taken during the battle. Instead, the environment informs the agent about the presence of novelty, requiring the agent to reason about what and where the novelty occurred. Consequently, detection poses no challenge for the baseline agent. A design decision is made for the agent to prioritize mission abortion if the no-fire-entity is in close proximity to the target, as the only viable outcome is to abort the mission entirely.

In the case of agent novelty, a friendly aircraft is destroyed by an enemy SAM after encountering the new missile range for the first time. Following one round of repair, the SAM missile range model is adjusted to accommodate a range of 3 cells. When multiple repairs $R$ yield identical inconsistency, the repair with the least model modification is selected. During execution, interesting emergent behavior is observed as the friendly aircraft occasionally enters the SAM missile range but evades destruction due to environmental stochasticity. Despite the SAM locking on and firing all missiles, there is a 50\% chance of the aircraft dodging the missiles successfully, while in the remaining 50\% of instances, one aircraft is destroyed while the other evades the missiles and proceeds to destroy the target. This results in an unrepairable inconsistency score $C$ because the model anticipates all aircraft surviving, whereas the observation records the loss of only one aircraft. Consequently, Equation \ref{eq:C} consistently yields $C = 10$, with 10 representing the penalty for a destroyed aircraft.

\begin{figure}[h]
    \centering
    \includegraphics[width=\columnwidth, trim={1cm 7cm 5cm 10cm},clip]{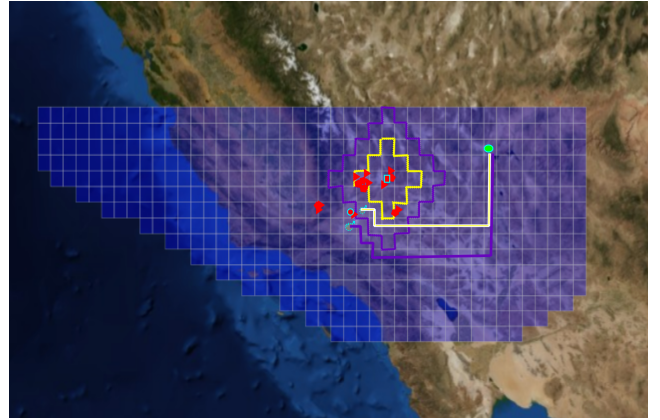}
    \caption{Agent Plan Before And After Novelty}
    \label{fig:grid_routes}
\end{figure}

As illustrated in Figure \ref{fig:grid_routes}, the agent successfully destroys the enemy target using a new route after repairing the PDDL+ model. The yellow solid line denotes the pre-novelty route, while the purple solid line represents the post-novelty route. The gray symbol of a bomb with dotted teal outlines marks the location where the friendly aircraft fired the missile from, with both aircraft returning to the home base after launching the missile.

In the context of relation novelty, HYDRA effectively identifies the SAM HP through inconsistency checking and adjusts the model to accommodate the novelty following the first battle where the novelty is introduced. It takes only one battle for HYDRA to observe and analyze the novelty, leading to successful model repair. An intriguing phenomenon to consider is the potential confusion caused by the goal state inconsistency. Any novelty that prevents the target from being destroyed without any friendly aircraft being destroyed as well could mislead the AI into believing that the SAM's survivability is high, as both scenarios result in an non-destroyed target.




In the case of environment, goal, and event novelty, the agent typically requires one battle to observe and comprehend the novelty, resulting in one false negative battle. However, it swiftly detects and adapts to the novelty in subsequent battles. Regarding environment novelty, the agent's inconsistency score tends to be lower compared to goal and event novelties. This discrepancy arises because no aircraft is ever destroyed due to environmental zones, which primarily lead to mission aborts resulting from communication loss, inability to lock on and fire missiles, fuel depletion, or missed targets. Conversely, goal and event novelties often result in aircraft destruction.

The number of battles required to find the correct repair for environment, goal, and event novelties depend on the location of the novelty is occurring in relation to the agent position. Each repair will be consistent with the observation if it finds an environment zone on the route of the aircraft from the last battle. However, given the extensive distance traveled by the aircraft from its starting position to the target, there exists a multitude of potential locations for the environment zone along its route. The above campaign shows the result of an environment zone found 3 cells from the starting position.

        

Therefore, the agent needs to 1) find a consistent model 2) remember the repair 3) determine if it is still inconsistent 4) find a new repair that is different from the perspective of the latest repair.


Goal and event novelties can sometimes be misclassified as agent novelty in this domain, as both scenarios lead to the destruction of friendly aircraft while the target survives. In many cases, since the aircraft is flying in a path relatively close to the SAM range, it often repairs the enemy SAM range, prompting the friendly aircraft to steer clear of areas near the current path, and unintentionally avoids the advanced enemy jets, resulting in the accommodation of the novelty based on a misclassification.



\section{Conclusion}
The presented prototype framework demonstrates the scalability of the HYDRA AI system from simpler physics-based game environments with hypothetical novelties to realistic high- and medium-fidelity simulators used for complex military scenarios, resembling real-world encounters. This represents a significant advancement towards deploying AI in dynamic real-world scenarios.
The study highlights the generality of the HYDRA agent approach. By incorporating domain knowledge through PDDL+, MMOs, and inconsistency checkers, the framework was expanded to accommodate multi-agent scenarios, complex environments with noise and stochasticity, and memory requirements. 
Future work on HYDRA will prioritize incorporating more expressive model changes and enhancing the ability to reason and adapt in real-time during mission execution, rather than post-engagement.



\begin{ack}
This research was sponsored by DARPA (HR001120C0040). The views contained in this document are those of the authors and should not be interpreted as representing the official policies, either expressed or implied, of DARPA or the U.S. government.
\end{ack}



\bibliography{mybibfile}

\end{document}